\documentclass{article}

\usepackage[preprint]{neurips_2020}
\usepackage[utf8]{inputenc} 
\usepackage[T1]{fontenc}    
\usepackage{hyperref}       
\usepackage{url}            
\usepackage{booktabs}       
\usepackage{amsfonts}       
\usepackage{nicefrac}       
\usepackage{microtype}      

\usepackage{amsmath}
\usepackage{amsthm}
\usepackage{amsfonts}
\usepackage{amssymb}
\usepackage{graphicx}
\usepackage{subcaption}
\usepackage{array}
\usepackage{float}
\usepackage{verbatim}

\usepackage{enumitem}

\usepackage{commands}

\title{Style is a Distribution of Features}

\author{%
  Eddie Huang\\
  Department of Computer Science\\
  University Illinois\\
  Urbana-Champaign, IL\\
  \texttt{ezhuang2@illinois.edu} \\
  \And
  Sahil Gupta\\
  Department of Computer Science\\
  University Illinois\\
  Urbana-Champaign, IL\\
  \texttt{sjgupta2@illinois.edu} \\
}

\begin{document}
\maketitle

\begin{abstract}
Neural style transfer (NST) is a powerful image generation technique that uses a convolutional neural network (CNN) to merge the content of one image with the style of another. 
Contemporary methods of NST use first or second order statistics of the CNN's features to achieve transfers with relatively little computational cost. 
However, these methods cannot fully extract the style from the CNN's features. We present a new algorithm for style transfer that fully extracts the style from the features by redefining the style loss as the Wasserstein distance between the distribution of features. Thus, we set a new standard in style transfer quality.
In addition, we state two important interpretations of NST. The first is a re-emphasis from \cite{li2017demystifying}, which states that style is simply the distribution of features. The second states that NST is a type of generative adversarial network (GAN) problem.
\end{abstract} \hspace{10pt}


\section{Introduction}

\begin{figure}[h]
\captionsetup[subfigure]{labelformat=empty}
\begin{tabular}{ccccccc}
\includegraphics[width=0.10\textwidth]{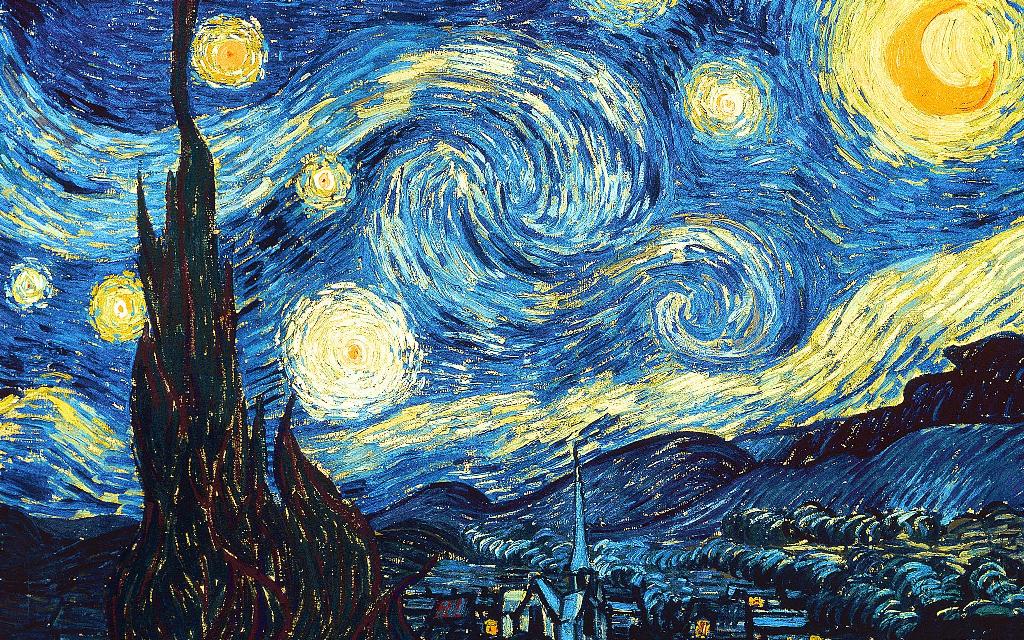} \\
BN Matching & \includegraphics[width=0.10\textwidth]{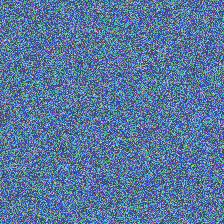} & \includegraphics[width=0.10\textwidth]{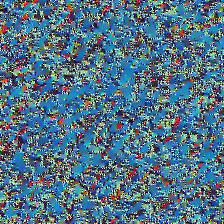} & \includegraphics[width=0.10\textwidth]{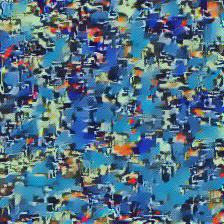} & \includegraphics[width=0.10\textwidth]{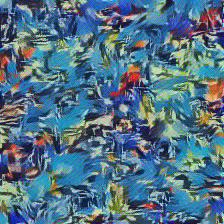} & \includegraphics[width=0.10\textwidth]{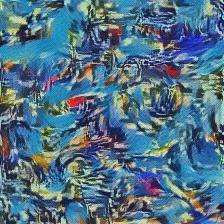} & \includegraphics[width=0.10\textwidth]{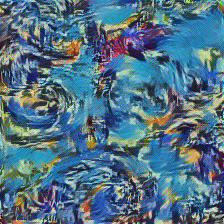} \\
Gatys et al. & \includegraphics[width=0.10\textwidth]{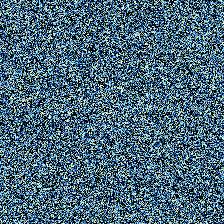} & \includegraphics[width=0.10\textwidth]{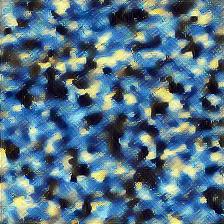} & \includegraphics[width=0.10\textwidth]{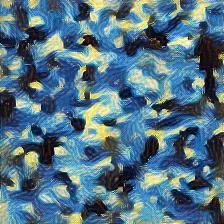} & \includegraphics[width=0.10\textwidth]{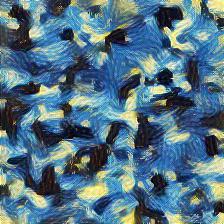} & \includegraphics[width=0.10\textwidth]{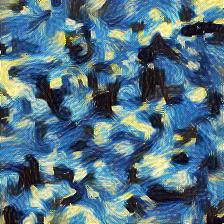} & \includegraphics[width=0.10\textwidth]{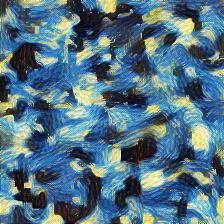} \\
Ours. & \includegraphics[width=0.10\textwidth]{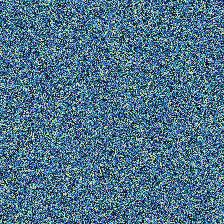} & \includegraphics[width=0.10\textwidth]{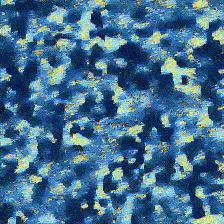} & \includegraphics[width=0.10\textwidth]{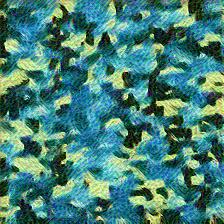} & \includegraphics[width=0.10\textwidth]{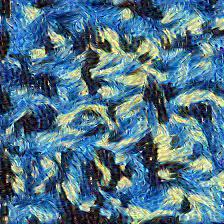} & \includegraphics[width=0.10\textwidth]{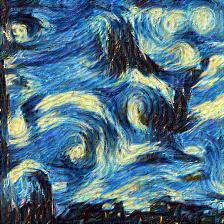} & \includegraphics[width=0.10\textwidth]{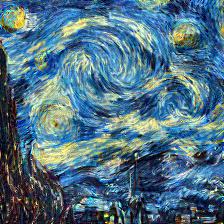}\\
& Raw Pixels & Layer 1 & Layer 2 & Layer 3 & Layer 4 & Layer 5
\end{tabular}
\caption{\small{Style representation of different methods. We compare 1\textsuperscript{st} and 2\textsuperscript{nd} order statistics (BN Matching and \cite{gatys2015neural}) to our method, which uses the Wasserstein metric. Starting from layer 2, 1\textsuperscript{st} and 2\textsuperscript{nd} order statistics fail to capture significantly higher level textures, unlike the Wasserstein metric.}}
\label{fig:starry-night}
\end{figure}

\subsection{Neural Style Transfer}
In 2015, \cite{gatys2015neural} introduced neural style transfer (NST), a powerful image generation technique that uses a convolutional neural network (CNN) to merge the content of one image with the style of another. There are many methods of style transfer nowadays, but since they are the first to introduce it, we refer to their method as the traditional style transfer. Content is defined as the semantic information of an image (i.e., the image is of a dog sitting on a porch), while style is defined as the textural information of an image (i.e. rough, smooth, colorful, etc.).

At a high level, the procedure for neural style transfer is to optimize the image we are generating (generated image) with respect to the content loss and style loss of the neural network. Content loss is defined simply as the element-wise difference between the corresponding feature maps of the content image and the generated image, and we usually define this difference with the mean squared error (MSE). Style loss on the other hand has many different interpretations. \cite{gatys2015neural} defined style loss as the difference between the Gramian matrix of the feature maps between the style image and the generated image, but it was unclear how this formula was derived.

For a while much of neural style transfer, specifically regarding the question "what is style?", remained a mystery. This led to the development of \cite{li2017demystifying} which showed that style can be interpreted simply as the \textit{distribution of features}, and that style loss is the distance between feature distributions of the style image and the generated image. However, contemporary methods still continue to use distribution matchings based on 1\textsuperscript{st} order or 2\textsuperscript{nd} order statistics. While they are fast and cheap, these methods are insufficient because they cannot fully discriminate between any two probability distributions. Thus, they cannot fully extract the style from the style image and transfer them to the generated image.

\subsection{Our contributions to NST}
Building off of the idea that style is the \textit{distribution} of features, we hypothesized that redefining style loss under the popular distribution distance metric, the Wasserstein distance, would be a superior alternative. The Wasserstein distance can always discriminate between two nonidentical probability distributions. Our experimental results show that performing Wasserstein style transfer achieves significantly more appealing results. The disadvantage is that it takes more computation compared to contemporary methods.

In the spirit of distribution matching, we also connect neural style transfer to the class of generative adversarial networks.

\section{Related work}
\subsection{Fast approximations}
Traditional style transfer is a slow and iterative optimization process, taking up to a few minutes to complete. Thus, there have been several works focusing on improving the speed of style transfer by approximating the style loss with cheaper statistical metrics.

\vspace{1em}
\noindent
\textbf{1\textsuperscript{st} order statistics} 
A popular method is to match the mean and standard deviation of the features. Examples include batch normalization statistics matching \cite{li2016revisiting, li2017demystifying}, instance normalization \cite{ulyanov2016instance, ulyanov2017improved}, conditional instance normalization \cite{dumoulin2016learned}, and adaptive instance normalization \cite{huang2017arbitrary, karras2018stylebased}.
All these methods can be classified as distribution matching using first order statistics.

Due to their speed, style transfer with 1\textsuperscript{st} order statistics is one of the most popular methods for commercial use. It has also found its way in other fields of machine learning. One notable example is \cite{karras2018stylebased}'s StyleGAN, a generative adversarial network that uses the adaptive instance normalization formulation of style to produce higher quality fake data.

\vspace{1em}
\noindent
\textbf{2\textsuperscript{nd} order statistics} Traditional style transfer (\cite{gatys2015neural}) uses 2\textsuperscript{nd} order statistics matching (see Section~\ref{section:background} for more details).
\cite{li2017universal} extends off of this idea by matching the covariance of the generated features with the style features. Similarly to \cite{huang2017arbitrary}, they take a pretrained autoencoder and perform linear transformations to match the features at various layers of the autoencoder. However instead of matching the mean and standard deviation of the features, \cite{li2017universal} matches the \textit{covariance} of the features.
Compared to \cite{gatys2015neural}, their algorithm produces similar results and is significantly faster. However, it is not as fast as the other 1\textsuperscript{st} order statistic methods.

\subsection{Deep generative modeling}
Other works that aim to improve the speed of style transfer use neural networks that can perform style transfer in one forward pass: \cite{johnson2016perceptual, ulyanov2016texture, li2016precomputed, dumoulin2016learned}. Such methods are even faster than the 1\textsuperscript{st} order statistics methods mentioned earlier, but the drawbacks are: 
(a) they are qualitatively worse and produce less diverse style transfers, and
(b) they specialize in a finite amount of styles, and cannot be applied to styles that they have not trained on. \cite{ulyanov2017improved} addressed the issue of quality and diversity by integrating the first order statistics methods, namely instance normalization, into their neural network architecture.

\subsection{Quality control}
There have also been works that aim to improve the quality and control over the style transfers. \cite{gatys2017controlling} extend their traditional style transfer algorithm to allow fine-grained control over the spatial location, color, and spatial scale. \cite{ulyanov2017improved} provides a method to improve the diversity of style transfers.

\cite{risser2017stable} stabilizes traditional style transfer by adding histogram losses. They also provide methods for localized control of style transfer using multi-resolution (pyramidal) image techniques.

While these methods do indeed improve style transfer quality, they do not solve the issue of fully extracting the style from the style image, which we address. \cite{jing2019neural} recently made a comprehensive review of style transfer methods, and they still regard the \cite{gatys2015neural}'s traditional method as the gold standard today.

\section{Background}\label{section:background}
\subsection{Traditional style transfer uses 2\textsuperscript{nd} order statistics}
Traditional style transfer uses the mean square error (MSE) of the Gramian matrix of the feature maps, which \cite{li2017demystifying} showed is mathematically equivalent to the Maximum Mean Discrepancy (MMD) of the features using the quadratic kernel $\kq(x,y) = (x^{\T}y)^2$. In other words, traditional style transfer matches the features using 2\textsuperscript{nd} order statistics.

Recall the following equivalent definitions of the MMD.
\begin{align}
    \MMD(p, q; k) &= \sup_{\norm{f}_\Hilbert}\left(\EE_x[f(x)] - \EE_y[f(y)]\right)\\
    &= \norm{\mu_p - \mu_q}
\end{align}
where $p,q$ are two arbitrary probability distributions. $f(x) = \inner{f, k(x, \cdot)}$ is the RKHS function for kernel $k$. $\mu_p = \EE_x[k(x,\cdot)], \mu_q = \EE_y[k(y,\cdot)]$ are the mean embeddings of $p,q$ respectively under the feature space $k(x, \cdot)$.

Furthermore, consider the definition of a quadratic kernel, $\kq(x,y) = (x^\T y)^2 = \inner{x, y}^2$. One can show that the corresponding feature map $\kq(x, \cdot)$ is equal to 
\begin{align}
    \kq(x, \cdot) = [x_n^2, \ldots, x_1^2, \sqrt{2} x_n x_{n-1}, \ldots, \sqrt{2} x_n x_1, \sqrt{2} x_{n-1} x_{n-2}, \ldots, \sqrt{2} x_{n-1} x_{1}, \ldots, \sqrt{2} x_{2} x_{1}] \label{eq:quad-phi}
\end{align}
Intuitively, the feature map represents the individual elements and all pair combinations of the elements. Thus, it is second order statistics. 

Nevertheless this second order matching is limited, and cannot discriminate conditional probabilities involving 2 or more other dimensions. An example of a limitation specific to the $\MMD$ under the quadratic kernel is that it is invariant to negation (i.e. $p \buildrel d \over = -q$). Consider this simple example of two probability distributions $p, q$ in $n$-dimensions. Let $p$ be concentrated at $\{1\}^n$ and $q$ be concentrated at $\{-1\}^n$. Then $\MMD(p, q;\kq) = 0$. Note that the MSE of the Gramian matrix between $p$ and $q$ would also be $0$, since they are mathematically equivalent. These two distributions are clearly different, yet the MMD cannot discriminate between them. For reference, the Wasserstein metric under the Euclidean distance, which we use in our method, would output $2\sqrt{n}$.

Furthermore, if you run \cite{gatys2015neural}'s style transfer on VGG19-BN (Batch Normalization), you will find that it performs significantly worse. We believe a major reason is because batch normalization standardizes features to have 0 mean. Therefore it's more likely that the style features and the generated features lie on opposite sides of the origin. Since MMD-Quad cannot distinguish between negated probability distributions, \cite{gatys2015neural}'s method fails.

\subsection{Wasserstein distance}
Intuitively, the Wasserstein loss can be interpreted as the minimum amount of work needed to move one distribution of mass to a target location. This is why it is sometimes known as the Earth-Mover distance. Let $p,q$ be two probability distributions.
We define the Wasserstein distance between them as
\begin{align}
    \W(p, q) = \inf_{\gamma\in \Gamma(p,q)}\EE_{x,y\sim\gamma}[d(x,y)]
\end{align}
where $\Gamma(p,q)$ is the set of all joint probability distributions over $p,q$ and $d$ is some distance function (e.g. Euclidean distance).

Unlike the 1\textsuperscript{st} or 2\textsuperscript{nd} order statistics, the Wasserstein metric can always discriminate between two nonidentical probability distributions (i.e. $\forall  p \ndeq q:\quad \W(p, q) > 0$). An easy way to see this is for any $\gamma\in \Gamma(p, q)$, there exists an $x,y$ such that $x\neq y$ and $\gamma(x, y) > 0$. Since $x\neq y$, $d(x,y) > 0$, which means $\inf_{\gamma\in \Gamma(p,q)}\EE_{x,y\sim\gamma}[d(x,y)] > 0$.

The most notable use of the Wasserstein distance in deep learning is by \cite{arjovsky2017wasserstein}, which used the Wasserstein distance as the generator loss in generative adversarial networks (GAN). In their work, which is now known as Wasserstein-GANs (W-GAN), the discriminator is a neural network that is trained to approximate the Wasserstein distance between the fake data distribution from the generator, and the real data distribution. \cite{arjovsky2017wasserstein} also provides an apt argument for why the Wasserstein distance in practice is superior to other distribution metrics including Jensen-Shannon divergence, Kullback–Leibler divergence, and Total Variation.

This approximation of the Wasserstein distance using neural networks was later improved by \cite{gulrajani2017improved}, which is now known as Wasserstein Gradient Penalty (Wasserstein-GP).
We employ \cite{gulrajani2017improved}'s method to approximate the Wasserstein distance between the features of the style image and the generated image.

\section{Methods}
The source code is available at: \url{https://github.com/aigagror/wasserstein-style-transfer}

\subsection{Training}

Let $\x^*, \x_c, \x_s$ represent the generated image, content image, and style image respectively.
Let the feature maps of $\x^*, \x_c, \x_s$ in layer $l$ of our CNN be denoted by $\Fl^l \in \R^{N_l\times M_l}, \Pl^l \in \R^{N_l\times M_l}, \Sl^l \in \R^{N_l\times M_l}$ respectively, where $N_l$ is the number of the feature maps in the layer $l$ and $M_l$ is the height times the width of the feature map. For this paper, let us call the columns of $\Fl^l,\Pl^l, \Sl^l$ as \textbf{features}. Intuitively, a feature can be interpreted as a ``pixel'' of the feature map.

In neural style transfer, we optimize the generated image with respect to the following loss

\begin{align}
  \Ll = \alpha \Ls + (1 - \alpha) \Lc \label{eq:style-transfer}  
\end{align}
where content loss $\Lc$ is defined as the mean squared error between the spatially corresponding features:
\begin{align}
  \Lc = \frac{1}{2}\sum_{i=1}^{N_l}\sum_{j=1}^{M_l}(\Fl^l_{ij} - \Pl^l_{ij})^2  
\end{align}
and style loss $\Ls$ is defined as a weighted sum of the distribution distances between the style's features and the generated image's features.

\begin{align}
  \Ls^l &= \D[\Fl^l, \Sl^l] & \textit{style layer loss}\\  
  \Ls &= \sum_l w_l \Ls^l & \textit{weighted sum of style layer losses}
\end{align}
where $\D$ is some distribution distance distance metric. In our case, it is the Wasserstein metric. Note that in traditional NST, the distribution distance metric is the Mean Maximum Dependency (MMD) with the quadratic kernel.

\subsection{Architecture}
\begin{figure}[ht]
    \centering
    \includegraphics[width=0.8\textwidth]{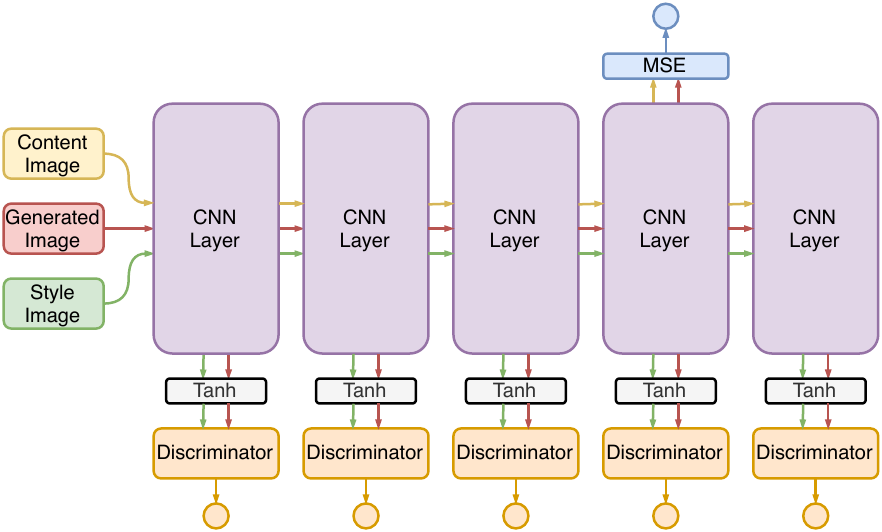}
    \caption{\small{Model architecture of the Wasserstein style transfer}}
    \label{fig:model}
\end{figure}

\subsection{Experimental Setup}
We test Wasserstein style transfer on VGG19 with batch normalization (VGG19-BN). We found that using VGG19 without batch-norm, which is the model other methods use, achieves the same results, but requires more training steps to converge. We use 5 distinct layers for style loss (conv1\_1, conv2\_1, conv3\_1, conv4\_1, conv5\_1), and 1 layer for the content loss (conv4\_1). These layers are also commonly used in other NST methods, which we purposefully chose for fair comparisons. To calculate the style loss using the Wasserstein distance, we attach a discriminator network at the corresponding CNN layer. These discriminators approximate the Wasserstein distance between the style features and the generated features under Wasserstein-GP. The discriminators each have 3 hidden layers, with dimensions 256, and a ReLU activation. Before the features are fed into the discriminators, they are first fed through an element-wise hyberbolic tangent (tanh) activation. We found this beneficial for keeping the training losses regularized and bounded. Figure \ref{fig:model} illustrates our model.

We compare our results to \cite{gatys2015neural}, \cite{huang2017arbitrary}, and \cite{ulyanov2017improved} using fine-tuned $\alpha$ values to achieve the same balance of content and style in their results. See the supplementary materials for more training details.

\section{Results} \label{section:results}

\subsection{Comparison to other NST methods}
\begin{figure}[ht]
\captionsetup[subfigure]{labelformat=empty}
\begin{tabular}{ccccc}
\includegraphics[width=0.16\textwidth]{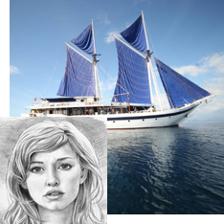} &
\includegraphics[width=0.16\textwidth]{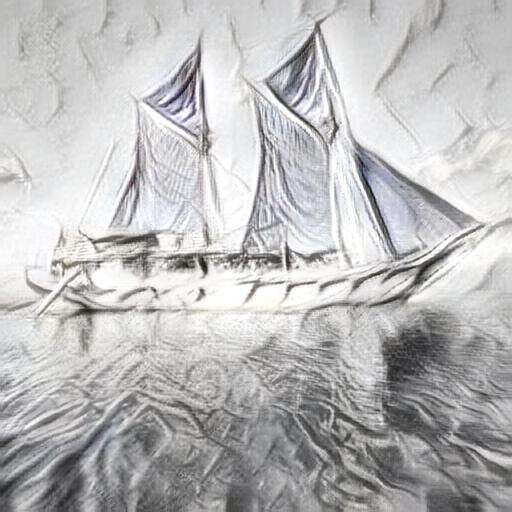} &
\includegraphics[width=0.16\textwidth]{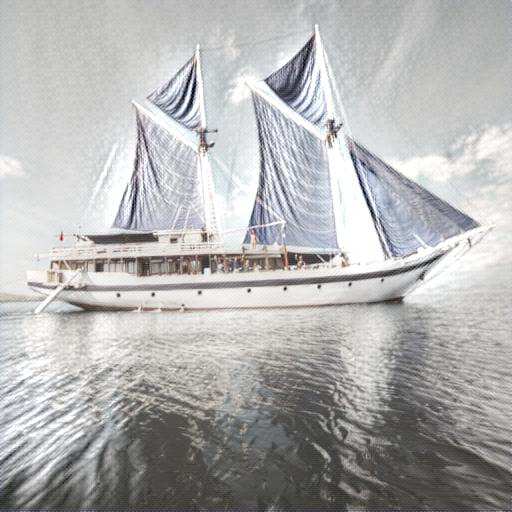} &
\includegraphics[width=0.16\textwidth]{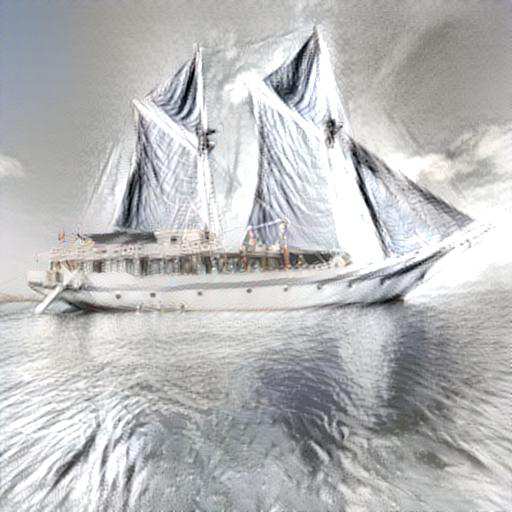} & \includegraphics[width=0.16\textwidth]{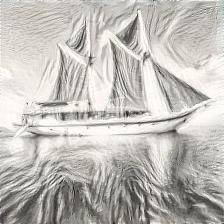}\\
\includegraphics[width=0.16\textwidth]{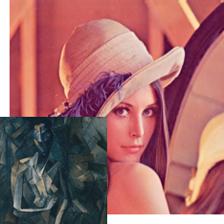} &
\includegraphics[width=0.16\textwidth]{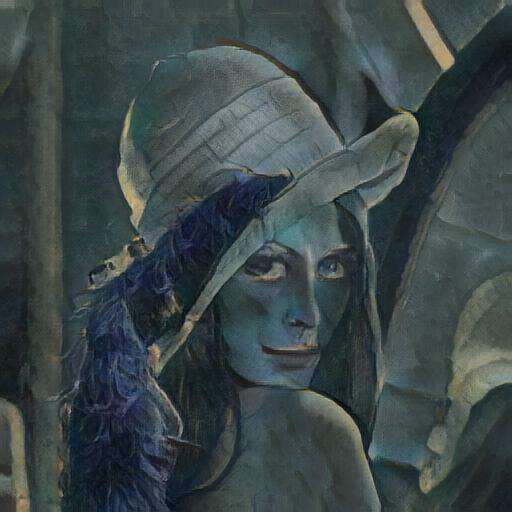} &
\includegraphics[width=0.16\textwidth]{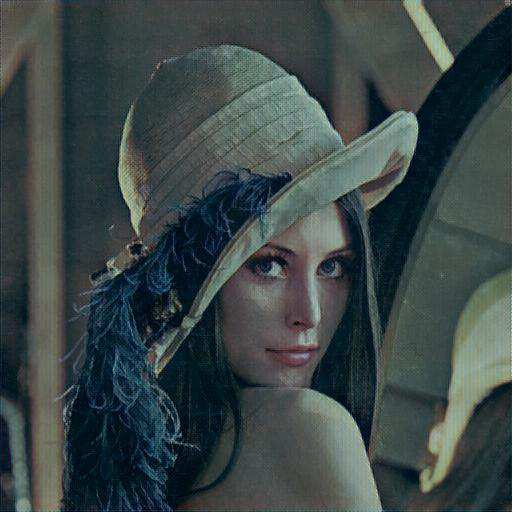} &
\includegraphics[width=0.16\textwidth]{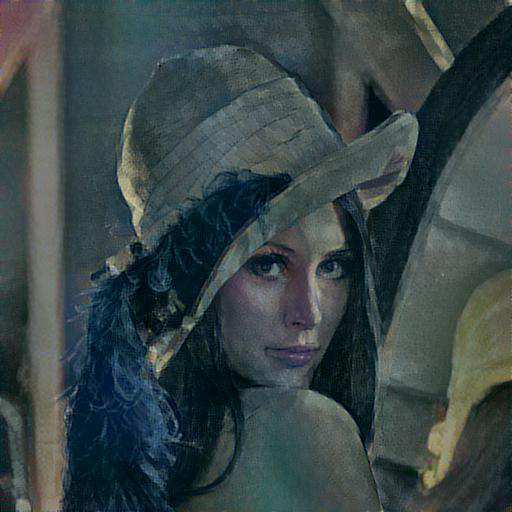} &
\includegraphics[width=0.16\textwidth]{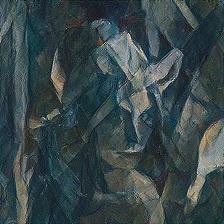}\\
\includegraphics[width=0.16\textwidth]{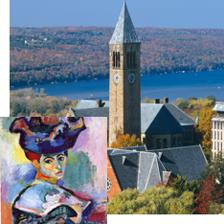} &
\includegraphics[width=0.16\textwidth]{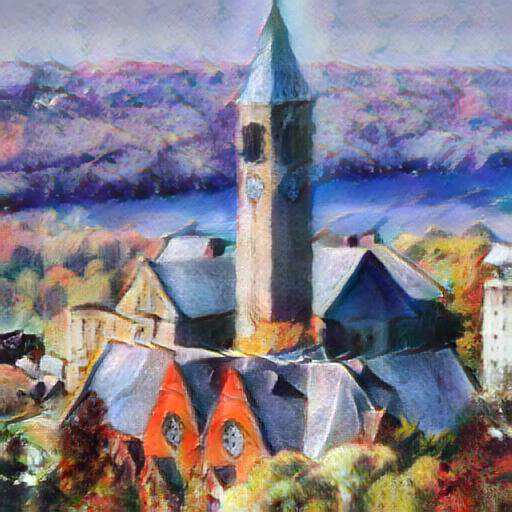} &
\includegraphics[width=0.16\textwidth]{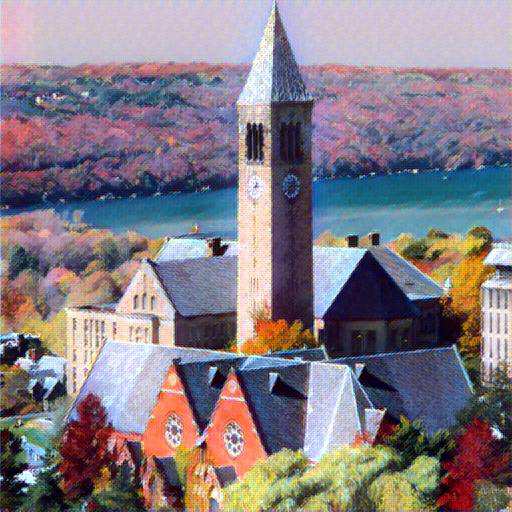} &
\includegraphics[width=0.16\textwidth]{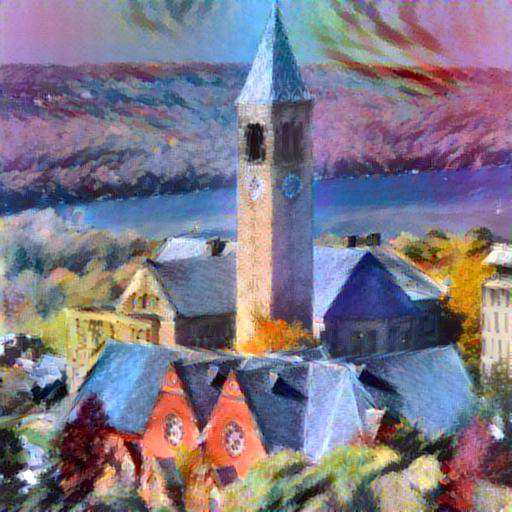} &
\includegraphics[width=0.16\textwidth]{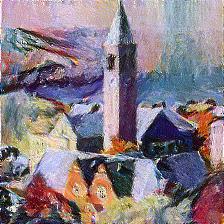}\\
\includegraphics[width=0.16\textwidth]{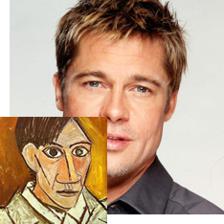} &
\includegraphics[width=0.16\textwidth]{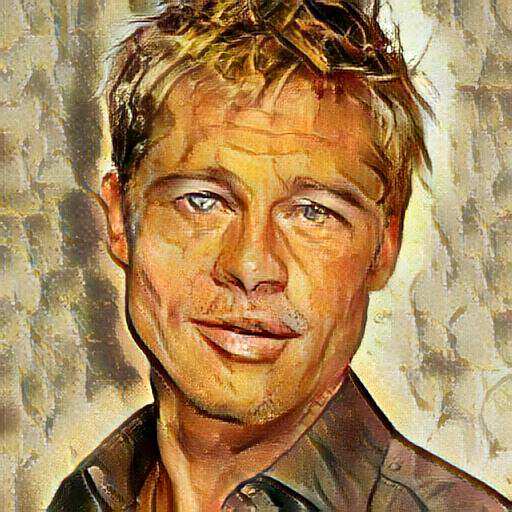} &
\includegraphics[width=0.16\textwidth]{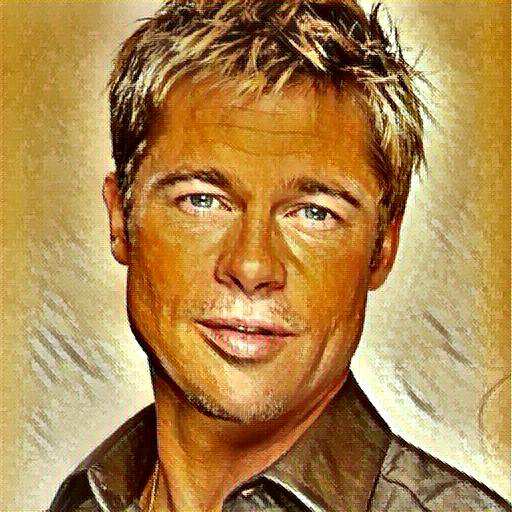} & \includegraphics[width=0.16\textwidth]{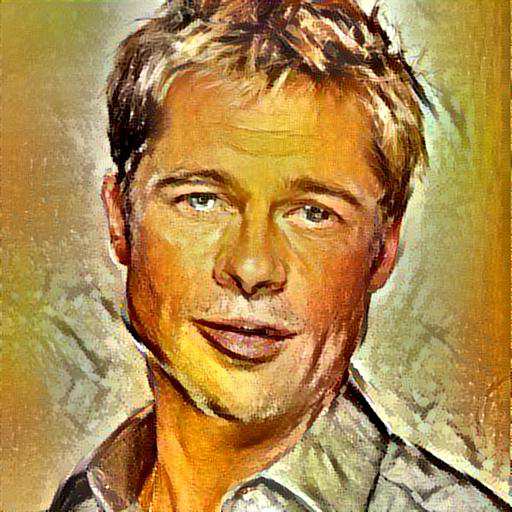} &
\includegraphics[width=0.16\textwidth]{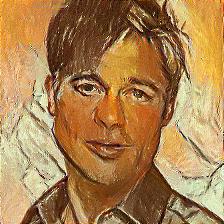}\\
\includegraphics[width=0.16\textwidth]{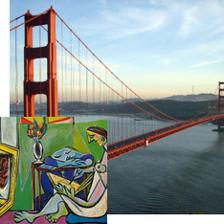} &
\includegraphics[width=0.16\textwidth]{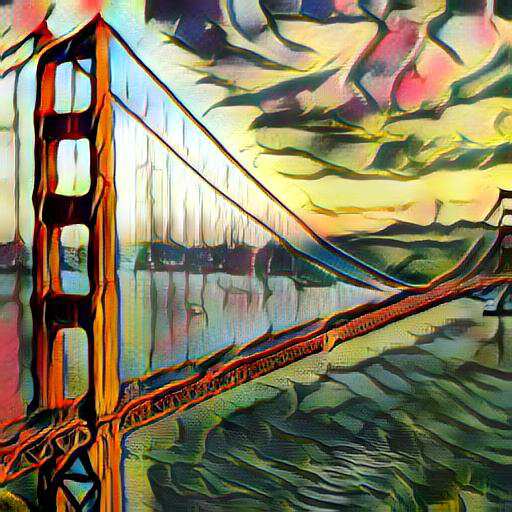} &
\includegraphics[width=0.16\textwidth]{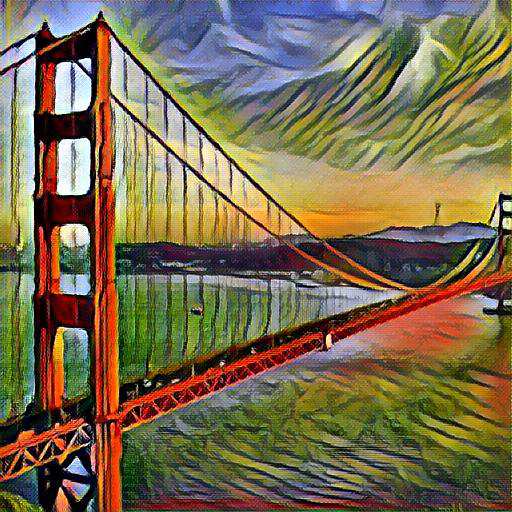} &
\includegraphics[width=0.16\textwidth]{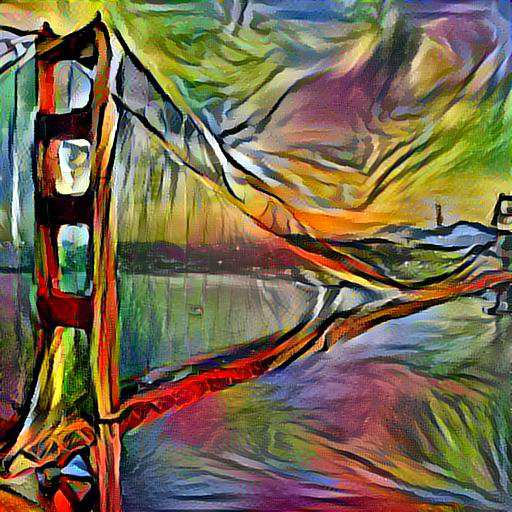} &
\includegraphics[width=0.16\textwidth]{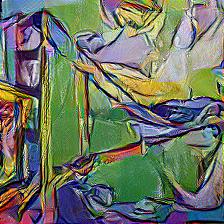}\\
Style-Content & Huang \textit{et al.} & Ulyanov \textit{et al.} & Gatys \textit{et al.} & Ours
\end{tabular}
\caption{\small{Example style transfer results. Our method was most preferred overall in an Amazon Turk survey.}}
\label{fig:comparisons}
\end{figure}

We run the our Wasserstein style transfer on the same style-content pairs as other contemporary methods for comparison. See Figure~\ref{fig:comparisons}. We think our transfers are of the best quality. Moreover, we think that it especially excels at capturing the color scheme and the higher level textures. For the sketch-sailboat pair (1\textsuperscript{st} row), we used the first 3 layers of VGG19, and for the picasso-brad-pitt pair (3rd row), we used the first 4 layers. This was done in order to prevent the semantic information of the style image from leaking into the generated image. 

We conducted a human preference survey using Amazon Turk using the images from Figure~\ref{fig:comparisons}. For each survey question, we presented the style and content image pair and titled them as "style" and "content" respectively. To clarify we presented the style and content images separately, unlike the style-content overlay in our figures, so that users can see the style image fully. We also presented transfers from all four methods with no labels for anonymity. Our survey question was the following:
\begin{quote}
    \textit{We merged the "style" and "content" of two images together using different methods. Select the merged image that best captures both the style and content.}
\end{quote}

We gathered 160 responses (32 responses for each of the 5 style-content pairs) from 47 unique users. Overall, our Wasserstein method was most preferred, with 59 votes, (36.9\% of all votes). \cite{ulyanov2017improved} was the next highest with 48 votes (30.0\%), followed by \cite{gatys2015neural} with 31 votes (19.4\%), and \cite{huang2017arbitrary} with 22 votes (13.8\%).

For each style-content pair, our method was most preferred, with the exception of the sketch-sailboat pair (first row of Figure~\ref{fig:comparisons}) in which \cite{ulyanov2017improved}'s method was most preferred.

\subsection{Style representation}
We compare style representations between 1\textsuperscript{st} order statistics, 2\textsuperscript{nd} order statistics, and the Wasserstein metric. See Figure~\ref{fig:starry-night}. For the 1\textsuperscript{st} order statistics, we use \cite{li2017demystifying}'s batch-normalization statistics matching (BN Matching) to align the mean and standard deviation of the features under the style layer loss $\Ls^l = (\EE[\Fl^l] - \EE[\Sl^l])^2 + (\sigma(\Fl^l) - \sigma(\Sl^l))^2$. For second order statistics, we use \cite{gatys2015neural}'s definition of the MSE of Gramian matrix. We also apply these style representations directly on the raw pixels themselves. The results are noisy images, but with the same color scheme as the original image. This is expected because optimizing over the raw pixels has no spatial dependencies whatsoever.

What's most interesting is how our method's style representation compares to the others. Starting from layer 3, it is clear that our method is better at capturing the higher level textures including the paint strokes, swirls, and even semantic features like the stars and the dark spires. In fact at layer 5, our method is so good at discriminating such high level features, it nearly recreates the global structure of the image.

\subsection{Semantic Information Leaking}
\begin{figure}[H]
    \centering
    \begin{tabular}{cccc}
    \includegraphics[width=0.15\textwidth]{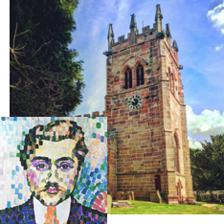} &
    \includegraphics[width=0.15\textwidth]{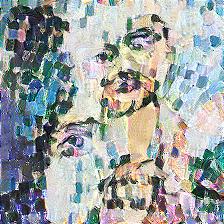} &
    \includegraphics[width=0.15\textwidth]{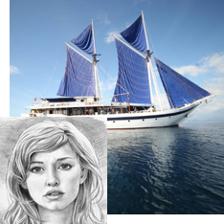} &
    \includegraphics[width=0.15\textwidth]{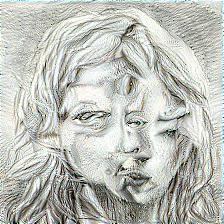} \\
    
    \includegraphics[width=0.15\textwidth]{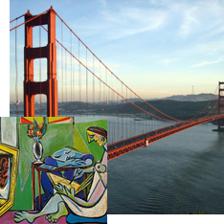} &
    \includegraphics[width=0.15\textwidth]{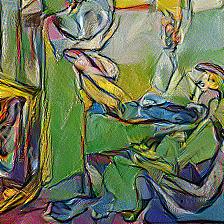} &
    \includegraphics[width=0.15\textwidth]{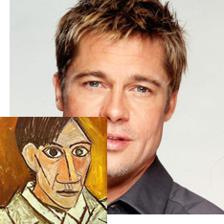} &
    \includegraphics[width=0.15\textwidth]{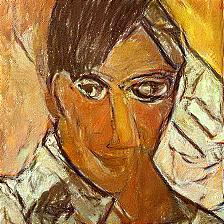} \\
    
    \end{tabular}
    \caption{\small{Our Wasserstein distance based style transfer method is prone to capturing the style ``too well''. This happens when we use layers that are too deep in the CNN model or $\alpha$ is too high. In these cases, we begin to blur the line between high level textures and semantic content.}}
    \label{fig:failure-cases-a}
\end{figure}

For the first time in style transfer, we encounter the issue of leaking semantic information from the style into the content. As stated previously, our method is exceptional at capturing higher level textures from the deeper layers of the CNN model. However, higher level textures have a larger spatial footprint, which ultimately leads to capturing the semantic information as well. See Figure~\ref{fig:failure-cases-a} for examples. To avoid this issue, one can simply use shallower CNN layers or decrease the $\alpha$ value. While most methods use the first 5 layers of VGG, in some of our transfers we had to restrict ourselves to the first 3 or 4 layers because the our method would otherwise add the semantic information of the style image to the generated image.

\section{Discussion}

\subsection{Style is a distribution of features}
In 2017, in an attempt to demystify NST, \cite{li2017demystifying} hypothesized that style is simply the distribution of features and that style transfer is nothing more than feature distribution matching.

Consider a feature in the first convolutional layer. It's value depends on the input image pixels in its receptive field. As we move to deeper layers, a feature's receptive field with respect to the input image grows linearly. Hence features in low level layers capture small spatial dependencies and features in deeper layers capture larger spatial dependencies. In layman's terms, features in low layers capture low-level textures like color, while features in higher layers capture higher-level textures like roughness or strokes. Since we define style as the \textit{distribution} of features, style can be informally described as the distribution of colors, roughness, smoothness, sharp edges, etc. Furthermore, since style is the distribution of features, it is global-spatially invariant. We find this idea that style is a distribution of features so appealing, we found it necessary to re-emphasize it here because we believe it is not discussed enough in the community. We hope you find this idea equally appealing.

As a consequence if we were to use just the raw pixels of the image as the features, the style representation should converge to any arbitrary image that has the same distribution of \textit{colors} of the style image. Figure \ref{fig:starry-night} supports this hypothesis.

\subsection{Style transfer is a special type of GAN}
\begin{figure}[t!]
\centering
\begin{subfigure}{0.4\textwidth}
\includegraphics[width=0.8\linewidth]{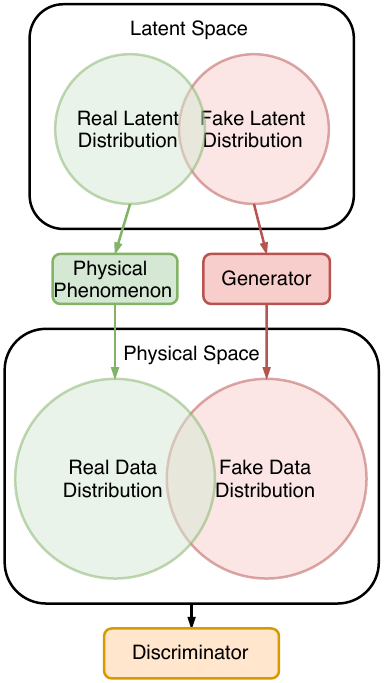} 
\caption{GAN Setting}
\label{fig:gan-setting}
\end{subfigure}
\hspace{25mm}
\begin{subfigure}{0.4\textwidth}
\includegraphics[width=0.8\linewidth]{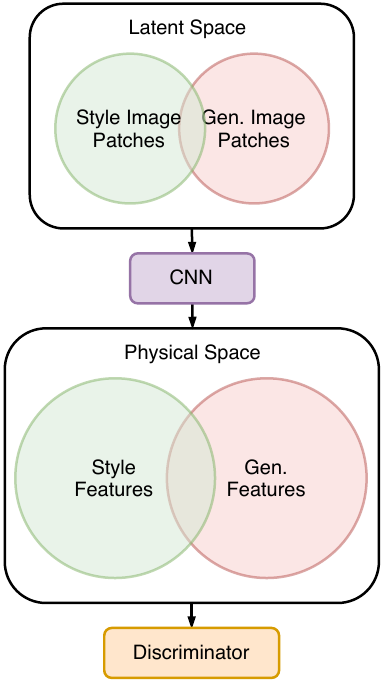}
\caption{NST Setting}
\label{fig:style-transfer-setting}
\end{subfigure}

\caption{\small{We argue that NST can be classified as a special type of GAN problem}}
\label{fig:gan-connection}
\end{figure}
In the spirit of defining style as a distribution of features, we interpret that NST can be classified as a type of GAN framework by the following connections. See Figure~\ref{fig:gan-connection} for illustration.

\begin{enumerate}[leftmargin=*]
    \item
    \begin{enumerate}[leftmargin=*]
        \item Under the traditional GAN setting, the discriminator and generator discriminate and align probability distributions in some physical space. The distribution of the real physical data is a result of some physical phenomenon that maps a distribution from some latent space to the physical space. The generator attempts to mimic this physical phenomenon with respect to the discriminator.
        
        \item Under the NST setting, the discriminator and generator discriminate and align probability distributions of the \textit{features} that are mapped by the CNN from either the style image (real latent distribution) or the generated image (fake latent distribution). More specifically, an element in the latent space is a patch of the image whose size is equal to the receptive field of the CNN layer. As a consequence, changing one pixel changes multiple latent elements since multiple patches share the same pixel. In this setting, the CNN is both the generator and the real physical phenomenon, mapping the latent space to the physical space.
    \end{enumerate}
    
    \item
    \begin{enumerate}[leftmargin=*]
        \item Under the traditional GAN setting, the generator is usually fed in some fixed latent distribution (e.g. a Gaussian distribution), and we optimize the generator network so that it's mapped output distribution aligns with the real data distribution. In essence, we make the generator try to mimic the physical phenomenon.
        
        \item Under the NST setting, this latent distribution are image patches, and instead of fixing this distribution, we fix the CNN and optimize over the the generated image patches. In essence, we try to align the distribution of image patches in the generated image with the style image. This aligns nicely with the idea that style transfer is a texture alignment process because image patches are locally spatial features.
    \end{enumerate}
\end{enumerate}

Since the Wasserstein metric has had great success with improving GANs, we can expect similar benefits in its application to style transfer.

\section{Conclusion}
We provide a new alternative to style transfer that calculates the style loss using the Wasserstein distance. If you consider style as the distribution of features, then the Wasserstein metric is more effective than contemporary methods at discriminating between the style features and the generated features, which leads to higher quality transfers. Thus we set a new benchmark for style transfer quality.

\begin{ack}
\begin{itemize}
    \item Thanks to Professor Rayadurgam Srikant for providing consultation on how to compare the Wasserstein distance against the MMD.
    \item Thanks to Hossein Talebi for general advice on the writing of this paper.
    \item Thanks to Bennett Ip for helping us with the initial developments of this project.
\end{itemize}
\end{ack}

\medskip

\newpage

\bibliographystyle{unsrtnat}
\bibliography{ref}

\newpage

\section{Supplementary Materials}

\subsection{Training Details}

\begin{table}[H]
    \centering
    \begin{tabular}{|c|c|}
        \hline
        Optimizer & Adam \\
        \hline
        Image Learning Rate & 2e-2 \\
        \hline
        Discriminator Learning Rate & 5e-4 \\
        \hline
        Training Steps & 500 \\
        \hline
        Discriminator Batch Size & 1024 \\
        \hline
    \end{tabular}
    \medskip
    \caption{Training Details}
    \label{tab:training-details}
\end{table}

\subsection{More examples}

\begin{figure}[H]
\centering
\captionsetup[subfigure]{labelformat=empty}
\begin{tabular}{cccc}
\includegraphics[width=0.2\textwidth]{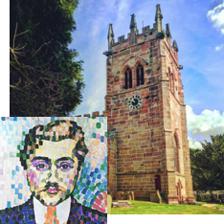} &
\includegraphics[width=0.2\textwidth]{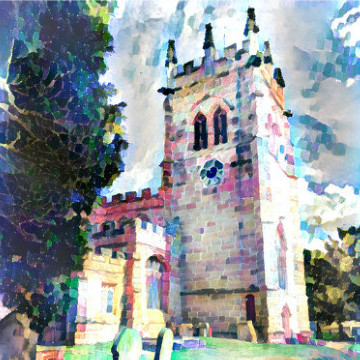} &
\includegraphics[width=0.2\textwidth]{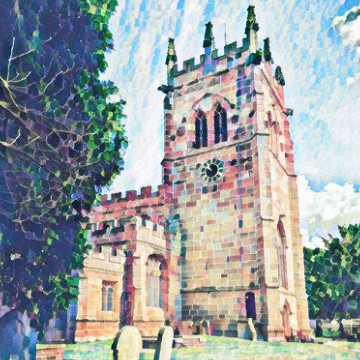} &
\includegraphics[width=0.2\textwidth]{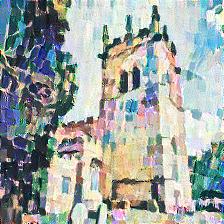} \\
\includegraphics[width=0.2\textwidth]{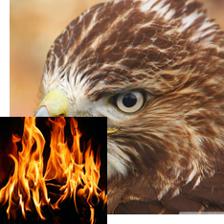} &
\includegraphics[width=0.2\textwidth]{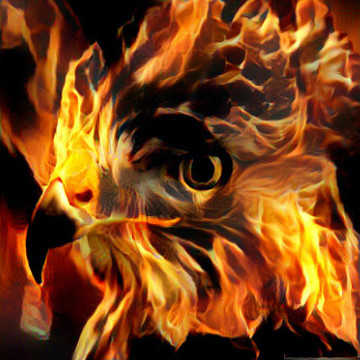} &
\includegraphics[width=0.2\textwidth]{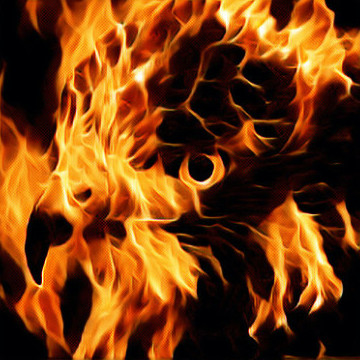} &
\includegraphics[width=0.2\textwidth]{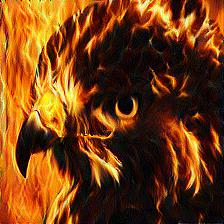}\\
Style-Content & Ulyanov \textit{et al.} & Gatys \textit{et al.} & Ours
\end{tabular}
\caption{Additional NST Comparisons with Gatys \textit{et. al.} and Ulyanov \textit{et. al.}}
\label{fig:teaser}
\end{figure}

\begin{figure}[H]
    \centering
    \captionsetup[subfigure]{labelformat=empty}
    \begin{tabular}{cccc}
    \includegraphics[width=0.2\textwidth]{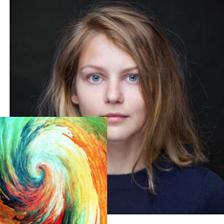} &
    \includegraphics[width=0.2\textwidth]{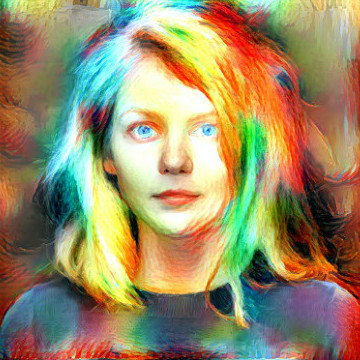} &
    \includegraphics[width=0.2\textwidth]{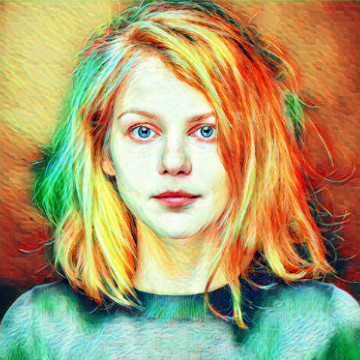} &
    \includegraphics[width=0.2\textwidth]{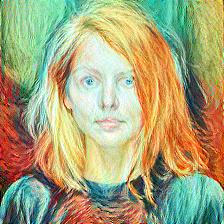} \\
    
    \includegraphics[width=0.2\textwidth]{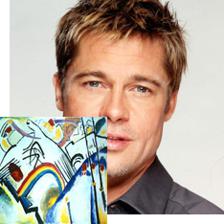} &
    \includegraphics[width=0.2\textwidth]{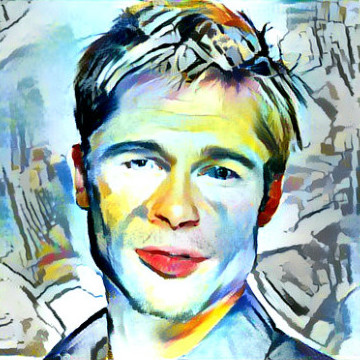} &
    \includegraphics[width=0.2\textwidth]{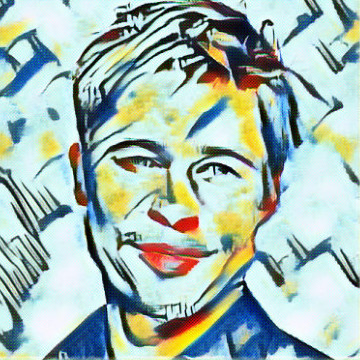} &
    \includegraphics[width=0.2\textwidth]{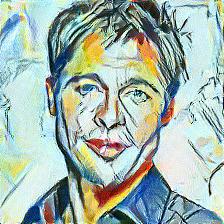} \\
    
    \includegraphics[width=0.2\textwidth]{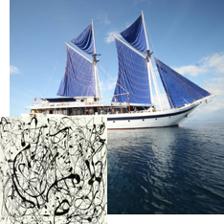} &
    \includegraphics[width=0.2\textwidth]{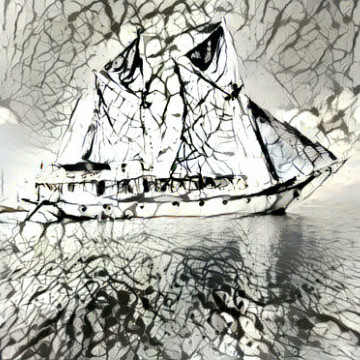} &
    \includegraphics[width=0.2\textwidth]{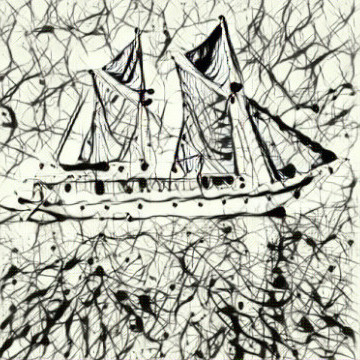} &
    \includegraphics[width=0.2\textwidth]{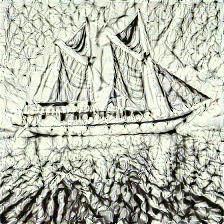}\\
    Style-Content & Ulyanov \textit{et al.} & Gatys \textit{et al.} & Ours
    \end{tabular}
    \caption{Additional NST Comparisons with Gatys \textit{et. al.} and Ulyanov \textit{et. al.}}
    \label{fig:addln-comp-gatys-ulyanov}
\end{figure}

\begin{figure}[H]
    \centering
    \captionsetup[subfigure]{labelformat=empty}
    \begin{tabular}{ccc}
    
    \includegraphics[width=0.2\textwidth]{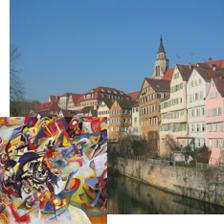} &
    \includegraphics[width=0.2\textwidth]{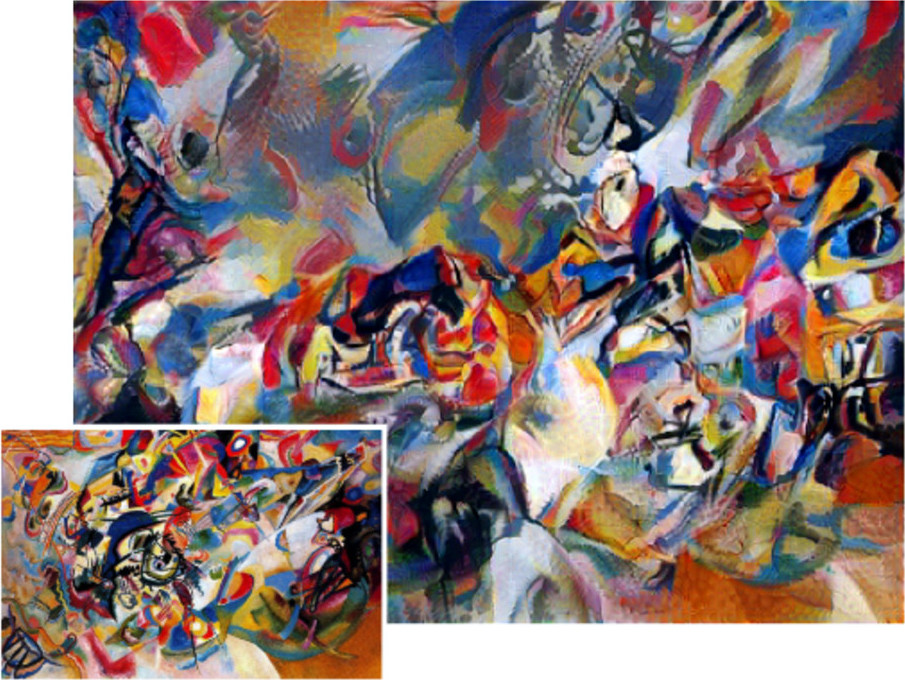} &
    \includegraphics[width=0.2\textwidth]{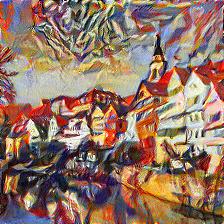} \\
    
    \includegraphics[width=0.2\textwidth]{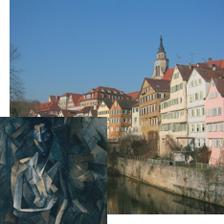} &
    \includegraphics[width=0.2\textwidth]{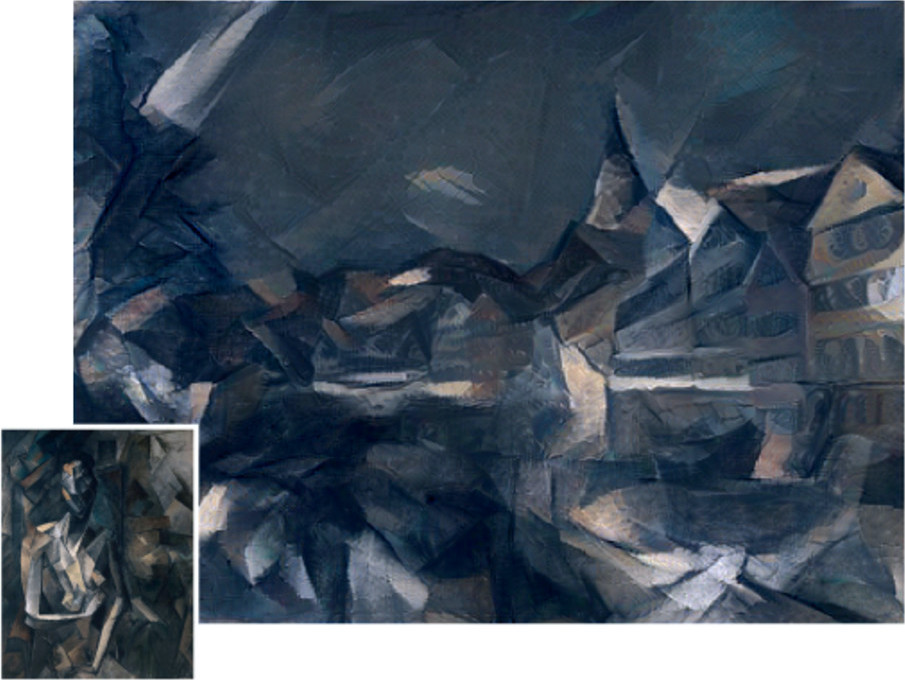} &
    \includegraphics[width=0.2\textwidth]{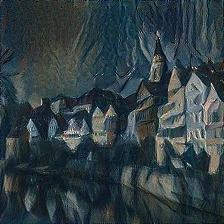} \\
    
    \includegraphics[width=0.2\textwidth]{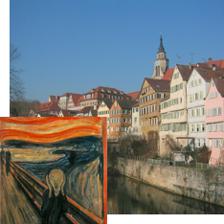} &
    \includegraphics[width=0.2\textwidth]{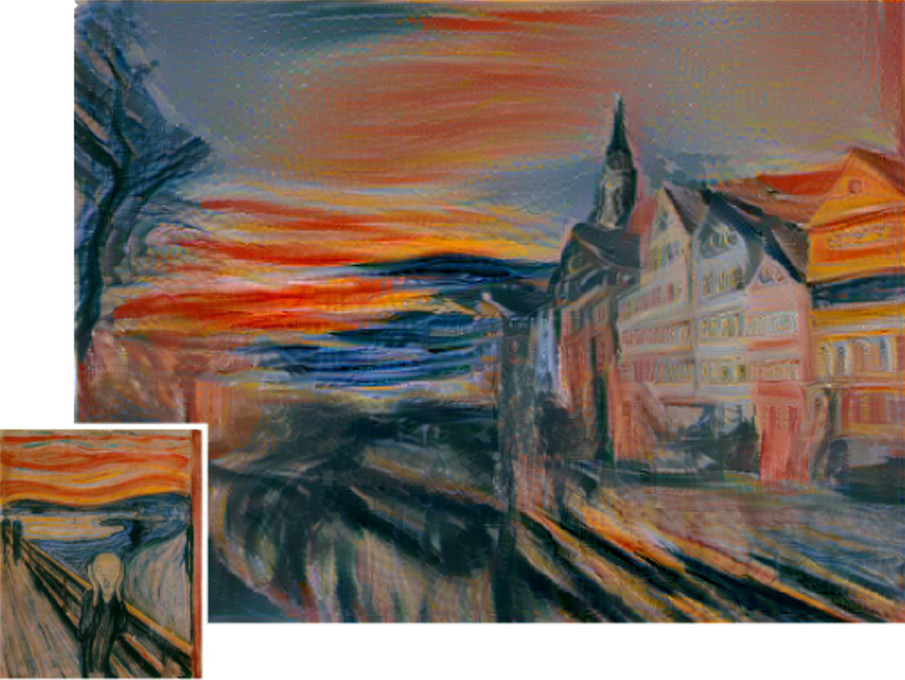} &
    \includegraphics[width=0.2\textwidth]{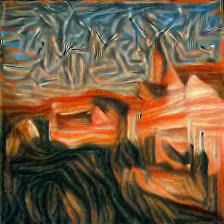} \\
    
    \includegraphics[width=0.2\textwidth]{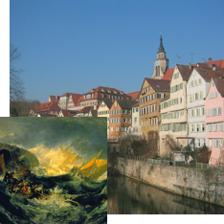} &
    \includegraphics[width=0.2\textwidth]{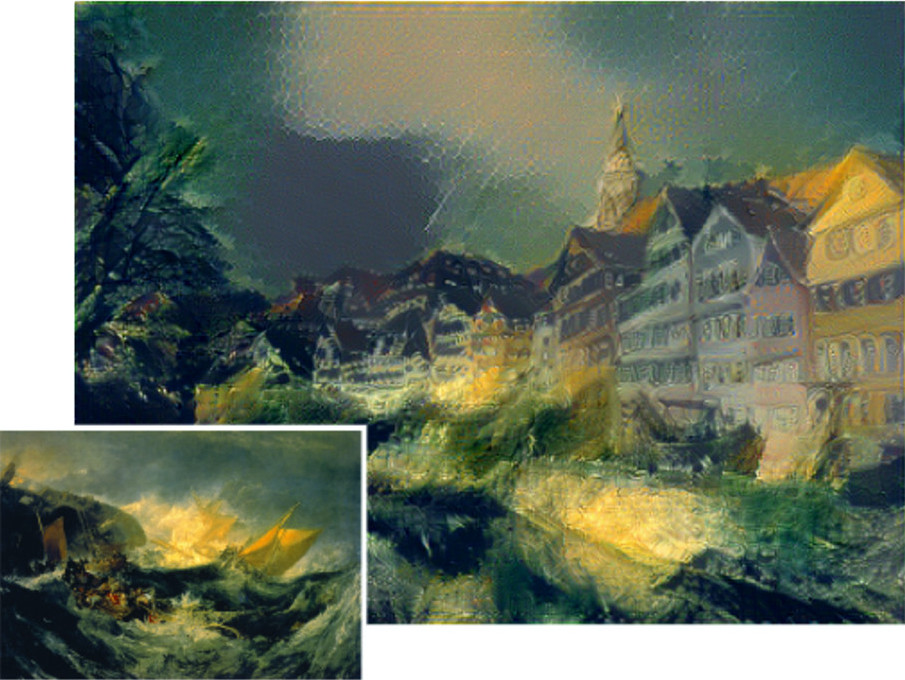} &
    \includegraphics[width=0.2\textwidth]{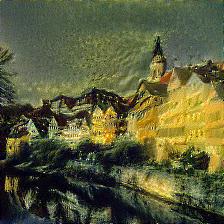} \\
    
    \includegraphics[width=0.2\textwidth]{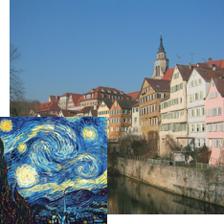} &
    \includegraphics[width=0.2\textwidth]{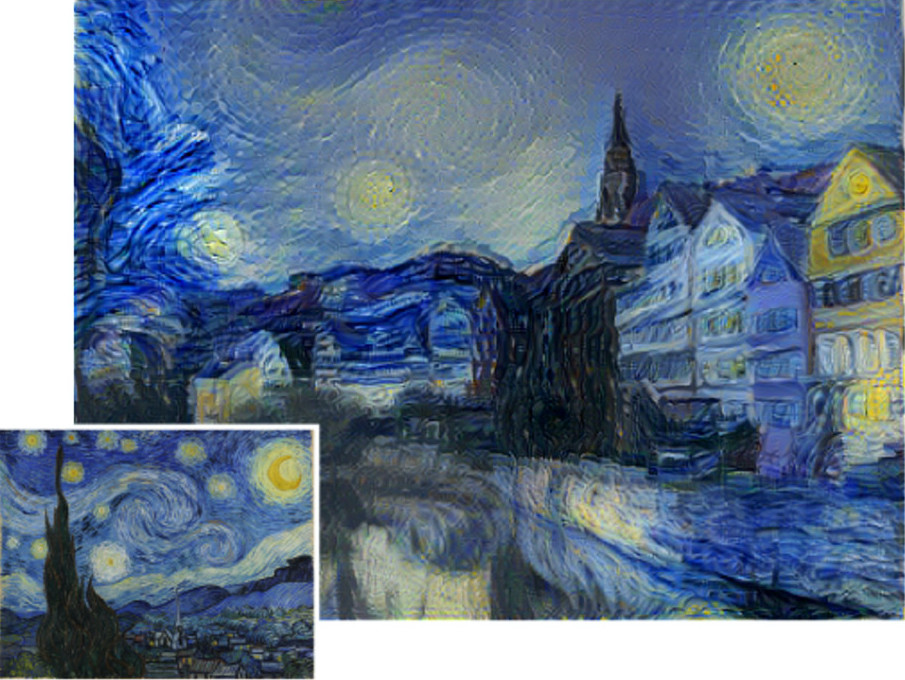} &
    \includegraphics[width=0.2\textwidth]{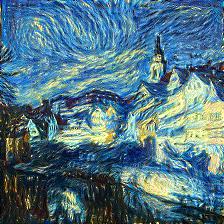}\\
    Style-Content & Gatys \textit{et al.} & Ours
    \end{tabular}
    \caption{Additional NST Comparisons with Gatys \textit{et. al.}}
    \label{fig:addln-comp-gatys}
\end{figure}

\begin{figure}[H]
    \centering
    \captionsetup[subfigure]{labelformat=empty}
    \begin{tabular}{cc}
    
    \includegraphics[width=0.4\textwidth]{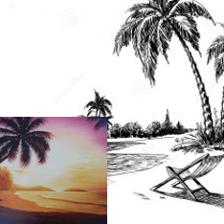} &
    \includegraphics[width=0.4\textwidth]{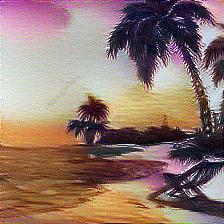} \\
    
    \includegraphics[width=0.4\textwidth]{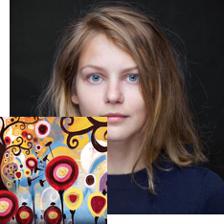} &
    \includegraphics[width=0.4\textwidth]{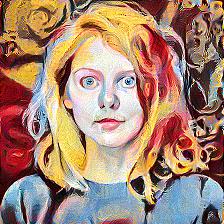} \\

    \includegraphics[width=0.4\textwidth]{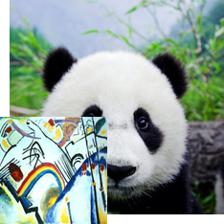} &
    \includegraphics[width=0.4\textwidth]{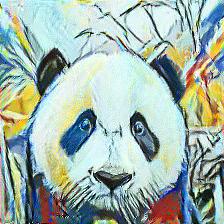}\\
    Style-Content & Ours
    
    \end{tabular}
    \caption{Additional Wasserstein Style Transfer Examples}
    \label{fig:addln-transfers}
\end{figure}

\begin{figure}[H]
    \captionsetup[subfigure]{labelformat=empty}
    \centering
    \begin{tabular}{cccc}
    \includegraphics[width=0.2\textwidth]{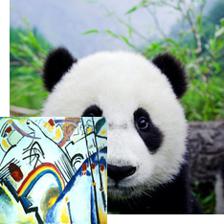} &
    \includegraphics[width=0.2\textwidth]{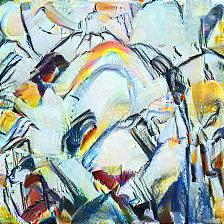} &
    \includegraphics[width=0.2\textwidth]{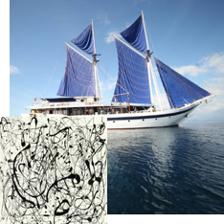} &
    \includegraphics[width=0.2\textwidth]{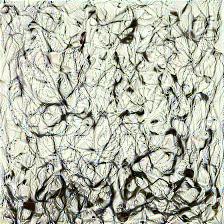}\\
    
    \includegraphics[width=0.2\textwidth]{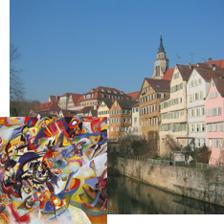} &
    \includegraphics[width=0.2\textwidth]{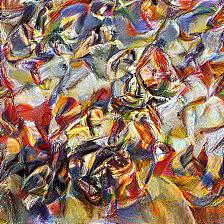} &
    \includegraphics[width=0.2\textwidth]{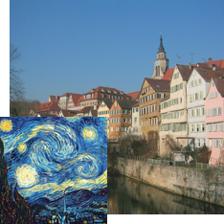} &
    \includegraphics[width=0.2\textwidth]{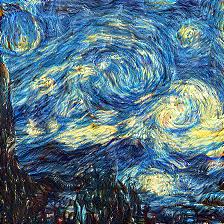}\\
    
    \includegraphics[width=0.2\textwidth]{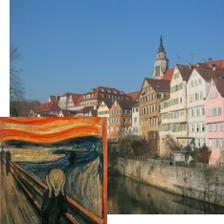} &
    \includegraphics[width=0.2\textwidth]{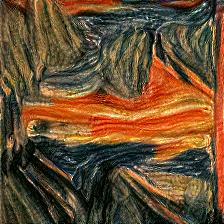} &
    \includegraphics[width=0.2\textwidth]{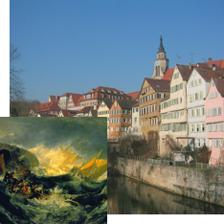} &
    \includegraphics[width=0.2\textwidth]{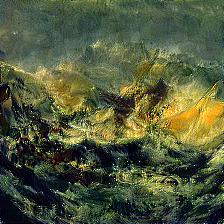}\\
    
    \includegraphics[width=0.2\textwidth]{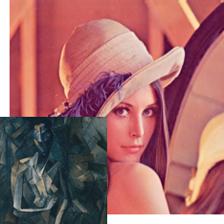} &
    \includegraphics[width=0.2\textwidth]{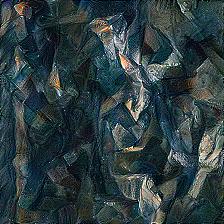} &
    \includegraphics[width=0.2\textwidth]{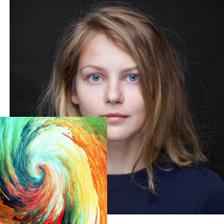} &
    \includegraphics[width=0.2\textwidth]{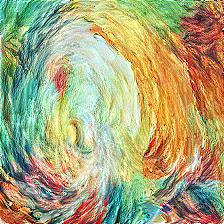}\\
    
    \includegraphics[width=0.2\textwidth]{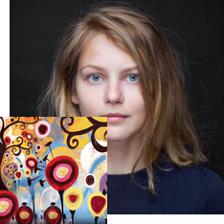} &
    \includegraphics[width=0.2\textwidth]{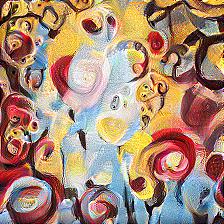} &
    \includegraphics[width=0.2\textwidth]{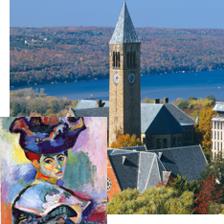} &
    \includegraphics[width=0.2\textwidth]{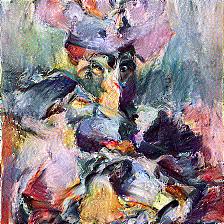}\\
    Style-Content & Ours & Style-Content & Ours
    \end{tabular}
    \caption{Additional Wasserstein style transfer semantic content leakage instances. All of the outputs used 5 layers of the pretrained CNN (VGG19-BN) with high $\alpha$ values (0.8-0.9)}
    \label{fig:addln-failure-cases-a}
\end{figure}

\end{document}